\documentclass{article}

\usepackage{microtype}
\usepackage{graphicx}
\usepackage{subfigure}
\usepackage{booktabs} 

\usepackage{hyperref}

\usepackage[accepted]{icml2018} 

\newcommand\titletext{Overcoming Catastrophic Forgetting with Hard Attention to the Task}
\icmltitlerunning{\titletext}

\usepackage{amsmath,amssymb,bm}
\usepackage{balance}

\usepackage{etoolbox}
\newtoggle{twodoc}
\togglefalse{twodoc}
\newtoggle{twodoc_body}
\toggletrue{twodoc_body}
\newcommand{\apx}[1]{\iftoggle{twodoc}{Supplementary Materials}{Appendix~\ref{#1}}}


\newcommand{\ab}[1]{\mathcal{#1}}

\newcommand{\ve}[1]{\textbf{#1}}
\newcommand{\se}[1]{\mathsf{#1}}

\usepackage{color}

\iftoggle{twodoc}{
}{
}

\begin{document} 

\iftoggle{twodoc}{
	\iftoggle{twodoc_body}{
		\twocolumn[
\icmltitle{\titletext}




\begin{icmlauthorlist}
\icmlauthor{Joan Serr\`a}{tef}
\icmlauthor{D\'idac Sur\'is}{tef,upc}
\icmlauthor{Marius Miron}{tef,upf}
\icmlauthor{Alexandros Karatzoglou}{tef}
\end{icmlauthorlist}

\icmlaffiliation{tef}{Telef\'onica Research, Barcelona, Spain}
\icmlaffiliation{upc}{Universitat Polit\`ecnica de Catalunya, Barcelona, Spain}
\icmlaffiliation{upf}{Universitat Pompeu Fabra, Barcelona, Spain}

\icmlcorrespondingauthor{Joan Serr\`a}{joan.serra@telefonica.com}

\icmlkeywords{catastrophic forgetting, lifelong learning, hard attention, multitask}

\vskip 0.3in
]


\printAffiliationsAndNotice{}  


\begin{abstract} 
Catastrophic forgetting occurs when a neural network loses the information learned in a previous task after training on subsequent tasks. This problem remains a hurdle for artificial intelligence systems with sequential learning capabilities. In this paper, we propose a task-based hard attention mechanism that preserves previous tasks' information without affecting the current task's learning. A hard attention mask is learned concurrently to every task, through stochastic gradient descent, and previous masks are exploited to condition such learning. We show that the proposed mechanism is effective for reducing catastrophic forgetting, cutting current rates by 45~to~80\%. We also show that it is robust to different hyperparameter choices, and that it offers a number of monitoring capabilities. The approach features the possibility to control both the stability and compactness of the learned knowledge, which we believe makes it also attractive for online learning or network compression applications.
\end{abstract} 


\section{Introduction}
\label{sec:Intro}

With the renewed interest in neural networks, old problems re-emerge, specially if the solution is still open. That is the case with the so-called catastrophic forgetting or catastrophic interference problem~\cite{McCloskey89PLM,Ratcliff90PR}. In essence, catastrophic forgetting corresponds to the tendency of a neural network to forget what it learned upon learning from new or different information. For instance, when a network is first trained to convergence on one task, and then trained on a second task, it forgets how to perform the first task. 

Overcoming catastrophic forgetting is an important step in the advancement towards more general artificial intelligence systems~\cite{Legg07MM}. Such systems should be able to seamlessly remember different tasks, and to learn them sequentially, following a lifelong learning paradigm~\cite{Thrun95RAS}. Apart from being more biologically plausible~\cite{Clegg98TCS}, there are many practical situations which require a sequential learning system~\citep[cf.][]{Thrun95RAS}. For instance, it may be unattainable for a robot to retrain from scratch its underlying model upon encountering a new object/task. After accumulating a large number of objects/tasks and their corresponding information, performing concurrent or multitask learning at scale may be too costly. 

Storing previous information and using it to retrain the model was among the earliest attempts to overcome catastrophic forgetting; a strategy named ``rehearsal''~\cite{Robins95CS}. The use of memory modules in this context has been a subject of research until today~\cite{Rebuffi2017CVPR,LopezPaz17NIPS}. However, due to efficiency and capacity constrains, memory-free approaches were also introduced, starting with what was termed as ``pseudo-rehearsal''~\cite{Robins95CS}. This approach has found some success in transfer learning situations where one needs to maintain a certain accuracy on the source task after learning the target task~\cite{Jung16ARXIV,Li17PAMI}. Within the pseudo-rehearsal category, we could also consider recent approaches that substitute the memory module by a generative network~\cite{Venkatesan17ARXIV,Shin17NIPS,Nguyen17ARXIV}. Besides the difficulty of training a generative network for a sequence of tasks or certain types of data, both rehearsal and pseudo-rehearsal approaches imply some form of concurrent learning, that is, having to re-process `old' instances for learning a new task.

The other popular strategy to overcome catastrophic forgetting is to reduce representational overlap~\cite{French91CCS}. This can be done at the output, intermediate, and also input levels~\cite{Gutsein15IJCNN,He18ICLR}. A clean way of doing that in a soft manner is through so-called ``structural regularization''~\cite{Zenke17ICML}, either present in the loss function~\cite{Kirkpatrick17PNAS,Zenke17ICML} or at a separate merging step~\cite{Lee17NIPS}. With these strategies, one seeks to prevent major changes in the weights that were important for previous tasks. Dedicating specific sub-parts of the network for each task is another way of reducing representational overlap~\cite{Rusu16ARXIV,Fernando17ARXIV,Yoon18ICLR}. The main trade-off in representational overlap is to effectively distribute the capacity of the network across tasks while maintaining important weights and reusing previous knowledge. 


In this paper, we propose a task-based hard attention mechanism that maintains the information from previous tasks without affecting the learning of a new task. Concurrently to learning a task, we also learn almost-binary attention vectors through gated task embeddings, using backpropagation and minibatch stochastic gradient descent (SGD). The attention vectors of previous tasks are used to define a mask and constrain the updates of the network's weights on current tasks. Since masks are almost binary, a portion of the weights remains static while the rest adapt to the new task. We call our approach hard attention to the task (HAT).
We evaluate HAT in the context of image classification, using what we believe is a high-standard evaluation protocol: we consider random sequences of 8~publicly-available data sets representing different tasks, and compare with a dozen of recent competitive approaches. We show favorable results in 4~different experimental setups, cutting current rates by 45 to 80\%. We also show robustness with respect to hyperparameters and illustrate a number of monitoring capabilities. We make our code publicly-available\footnote{\url{https://github.com/joansj/hat}}.



\section{Putting Hard Attention to the Task}
\label{sec:Method}

\subsection{Motivation}
\label{sec:Method_Motiv}

The primary observation that drives the proposed approach is that the task definition or, more pragmatically, its identifier, is crucial for the operation of the network. Consider the task of discriminating between bird and dog images. When training the network to do so, it may learn some set of intermediate features. If the second task is to discriminate between brown and black animals using the same data (assuming it only contained birds and dogs that were either brown or black), the network may learn a new set of features, some of them with not much overlap with the first ones. Thus, if training data is the same in both tasks, one important difference should be the task description or identifier. Our intention is to learn to use the task identifier to condition every layer, and to later exploit this learned conditioning to prevent forgetting previous tasks. 

\subsection{Architecture}
\label{sec:Method_Arch}

To condition to the current task $t$, we employ a layer-wise attention mechanism (Fig.~\ref{fig:blockdiag}). Given the output of the units\footnote{In the remaining of the paper, we will use `units' to refer to both linear units (or fully-connected neurons) and convolutional filters. HAT can be extended to other parametric layers.} of layer $l$, $\ve{h}_l$, we element-wise multiply $\ve{h}^\prime_l = \ve{a}^t_l \odot \ve{h}_l$. 
However, an important difference with common attention mechanisms is that, instead of forming a probability distribution, $\ve{a}^t_l$ is a gated version of a single-layer task embedding $\ve{e}^t_l$,
\begin{equation}
	\ve{a}^t_l = \sigma\left( s \ve{e}^t_l \right) ,
    \label{eq:gate}
\end{equation}
where $\sigma(x) \in [0,1]$ is a gate function and $s$ is a positive scaling parameter. We use a sigmoid gate in our experiments, but note that other gating mechanisms could be used. All layers $l=1,\dots L-1$ operate equally except the last one, layer $L$, where $\ve{a}^t_L$ is binary hard-coded. The operation of layer $L$ is equivalent to a multi-head output~\cite{Bakker02JMLR}, which is routinely employed in the context of catastrophic forgetting~\citep[for example][]{Rusu16ARXIV,Li17PAMI,Nguyen17ARXIV}.

\begin{figure}[t]
	\begin{center}
	\includegraphics[width=1\linewidth]{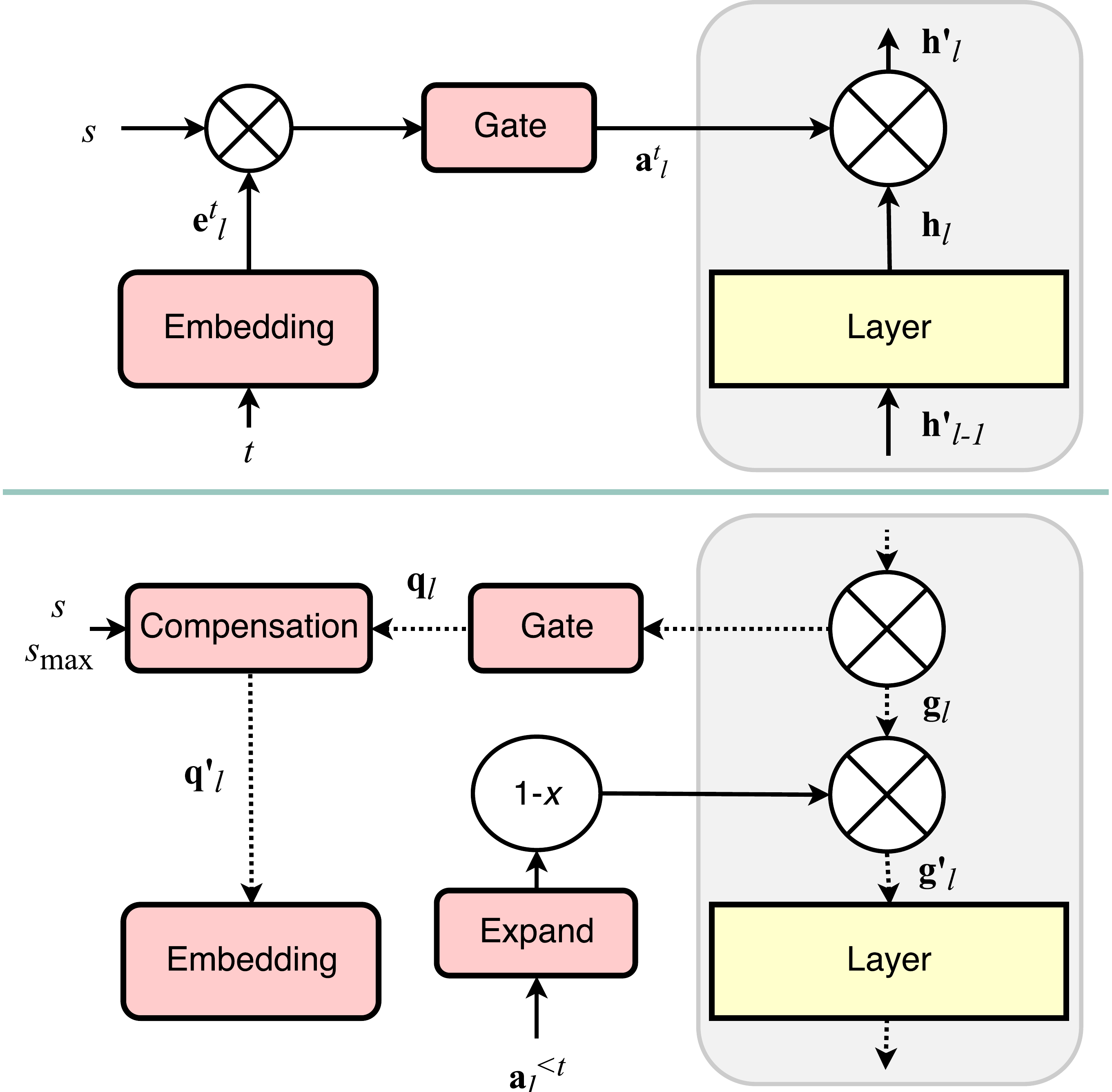}
    \vskip -0.1in
	\caption{Schematic diagram of the proposed approach: forward (top) and backward (bottom) passes.}
	\label{fig:blockdiag}
	\end{center}
	\vskip -0.1in
\end{figure} 

The idea behind the gating mechanism of Eq.~\ref{eq:gate} is to form hard, possibly binary attention masks which, act as ``inhibitory synapses''~\cite{McCulloch43TBMB}, and can thus activate or deactivate the output of the units of every layer. In this way, and similar to PathNet~\cite{Fernando17ARXIV}, we dynamically create and destroy paths across layers that can be later preserved when learning a new task. However, unlike PathNet, the paths in HAT are not based on modules, but on single units. Therefore, we do not need to pre-assign a module size nor to set a maximum number of modules per task. Given some network architecture, HAT learns and automatically dimensions individual-unit paths, which ultimately affect individual layer weights. Furthermore, instead of learning paths in a separate stage using genetic algorithms, HAT learns them together with the rest of the network, using backpropagation and SGD.

\subsection{Network Training}
\label{sec:Method_NetTrain}

To preserve the information learned in previous tasks upon learning a new task, we condition the gradients according to the cumulative attention from all the previous tasks. To obtain a cumulative attention vector, after learning task $t$ and obtaining $\ve{a}^t_l$, we recursively compute
\begin{equation*}
	\ve{a}^{\leq t}_l = \max\left( \ve{a}^t_l , \ve{a}^{\leq t-1}_l \right) ,
\end{equation*}
using element-wise maximum and the all-zero vector for $\ve{a}^{\leq 0}_l$. 
This preserves the attention values for units that were important for previous tasks, allowing them to condition the training of future tasks.

To condition the training of task $t+1$, we modify the gradient $g_{l,ij}$ at layer $l$ with the reverse of the minimum of the cumulative attention in the current and previous layers:
\begin{equation}
	g^\prime_{l,ij} = \left[ 1- \min\left( a^{\leq t}_{l,i} , a^{\leq t}_{l-1,j} \right) \right] g_{l,ij} ,
    \label{eq:gradmask}
\end{equation}
where the unit indices $i$ and $j$ correspond to the output ($l$) and input ($l-1$) layers, respectively. In other words, we expand the vectors $\ve{a}^{\leq t}_l$ and $\ve{a}^{\leq t}_{l-1}$ to match the dimensions of the gradient tensor of layer $l$, and then perform a element-wise minimum, subtraction, and multiplication (Fig.~\ref{fig:blockdiag}). We do not compute any attention over the input data if this consists of complex signals like images or audio. However, in the case such data consisted of separate or independent features, one could also consider them as the output of some layer and apply the same methodology.

Note that, with Eq.~\ref{eq:gradmask}, we create masks to prevent large updates to the weights that were important for previous tasks. This is similar to the approach of PackNet~\cite{Mallya17ARXIV}, which was made public during the development of HAT. In PackNet, after an heuristic selection and retraining, a binary mask is found and later applied to freeze the corresponding network weights. In this regard, HAT differs from PackNet in three important aspects. 
Firstly, our mask is unit-based, with weight-based masks automatically derived from those. 
Therefore, HAT also stores and maintains a lightweight structure.
Secondly, our mask is learned, instead of heuristically- or rule-driven. Therefore, HAT does not need to pre-assign compression ratios nor to determine parameter importance through a post-training step. 
Thirdly, our mask is not necessarily binary, allowing intermediate values between 0 and 1. This can be useful if we want to reuse weights for learning other tasks, at the expense of some forgetting, or we want to work in a more online mode, forgetting the oldest tasks to remember new ones. 

\subsection{Hard Attention Training}
\label{sec:Method_AttTrain}

To obtain a totally binary attention vector $\ve{a}^t_l$, one could use a unit step function as gate. However, since we want to train the embeddings $\ve{e}^t_l$ with backpropagation (Fig.~\ref{fig:blockdiag}), we prefer a differentiable function. To construct a pseudo-step function that allows the gradient to flow, we use a sigmoid with a positive scaling parameter $s$ (Eq.~\ref{eq:gate}). This scaling is introduced to control the polarization, or `hardness', of the pseudo-step function and the resulting output $\ve{a}^t_l$. Our strategy is to anneal $s$ during training, inducing a gradient flow, and set $s=s_{\max}$ during testing, using $s_{\max} \gg 1$ such that Eq.~\ref{eq:gate} approximates a unit step function. Notice that when $s \to \infty$ we get $a^t_{l,i} \to \{0,1\}$, and that when $s \to 0$ we get $a^t_{l,i} \to 1/2$. We will use the latter to start a training epoch with all network units being equally active, and progressively polarize them within the epoch.

During a training epoch, we incrementally linearly anneal the value of $s$ by
\begin{equation}
	s = \frac{1}{s_{\max}} + \left(s_{\max} - \frac{1}{s_{\max}} \right) \frac{b-1}{B-1} ,
    \label{eq:anneal}
\end{equation}
where $b=1,\dots B$ is the batch index and $B$ is the total number of batches in an epoch. The hyperparameter $s_{\max}\geq 1$ controls the stability of the learned tasks or, in other words the plasticity of the network's units. If $s_{\max}$ is close to 1, the gating mechanism operates like a regular sigmoid function, without particularly enforcing the binarization of $\ve{a}^t_l$. This provides plasticity to the units, with the model being able to forget previous tasks at the backpropagation stage (Sec.~\ref{sec:Method_NetTrain}). If, alternatively, $s_{\max}$ is a larger number, the gating mechanism starts operating as a unit step function. This provides stability with regard to previously learned tasks, preventing changes in the corresponding weights at the backpropagation stage.

\subsection{Embedding Gradient Compensation}
\label{sec:Method_Compens}

In preliminary analysis, we empirically observed that embeddings $\ve{e}^t_l$ were not changing much, and that the magnitude of the gradient was weak on those weights. After some investigation, we realized that the major part of the problem was due to the introduced annealing scheme (Eq.~\ref{eq:anneal}). To illustrate the effect of the annealing scheme on the gradients of $\ve{e}^t_l$, consider a uniformly distributed embedding $e^t_{l,i}$ across the active range of a standard sigmoid, $e^t_{l,i}\in [-6,6]$. If we do not perform any annealing and set $s=1$, we obtain a cumulative gradient after one epoch that has a bell-like shape and spans the whole sigmoid range (Fig.~\ref{fig:compensation}). Contrastingly, if we set $s=s_{\max}$, we obtain a much larger magnitude, but in a much lower range ($e^t_{l,i}\in [-1,1]$ in Fig.~\ref{fig:compensation}). The annealed version of $s$ yields a distribution in-between, with a lower range than $s=1$ and a lower magnitude than $s=s_{\max}$. A desirable situation would be to have a wide range, ideally spanning the range of $s=1$, and a large cumulative magnitude, ideally proportional to the one in the active region when $s=s_{\max}$. To achieve that, we apply a gradient compensation before updating $\ve{e}^t_l$.

\begin{figure}[t]
	\begin{center}
	\includegraphics{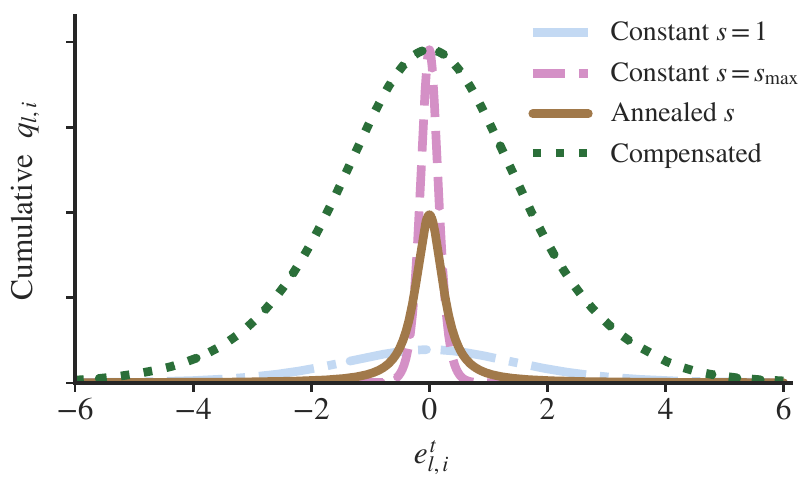}
    \vskip -0.15in
	\caption{Illustration of the effect that annealing $s$ has on the gradients $q$ of $\ve{e}^t$.}
	\label{fig:compensation}
	\end{center}
	\vskip -0.1in
\end{figure} 

In essence, the idea of the embedding gradient compensation is to remove the effects of the annealed sigmoid and to artificially impose the desired range and magnitude motivated in the previous paragraph. To do so, we divide the gradient $q_{l,i}$ by the derivative of the annealed sigmoid, and multiply by the desired compensation,
\begin{equation*}
	q^\prime_{l,i} = \frac{ s_{\max} \sigma\left(e^t_{l,i}\right) \left[ 1-\sigma\left(e^t_{l,i}\right) \right] }{ s \sigma\left(se^t_{l,i}\right) \left[ 1-\sigma\left(se^t_{l,i}\right) \right] } ~ q_{l,i} ,
\end{equation*}
which, after operating, yields
\begin{equation*}
	q^\prime_{l,i} = \frac{ s_{\max} \left[ \cosh\left(se^t_{l,i}\right)+1 \right] }{ s \left[ \cosh\left(e^t_{l,i}\right)+1 \right] } ~ q_{l,i} .
\end{equation*}
For numerical stability, we clamp $\vert se^t_{l,i} \vert \leq 50$ and constrain $e^t_{l,i}$ to remain within the active range of the standard sigmoid, $e^t_{l,i} \in [-6,6]$. In any case, however, $q_{l,i} \to 0$ when we hit those limits. That is, we are in the constant regions of the pseudo-step function. Notice also that, by Eq.~\ref{eq:anneal}, the minimum $s$ is never equal to 0.

\subsection{Promoting Low Capacity Usage}
\label{sec:Method_Compress}

It is important to realize that the hard attention values $a^t_{l,i}$ that are `active', that is, $a^t_{l,i}\to 1$, directly determine the units that will be dedicated to task $t$. Therefore, in order to have some model capacity reserved for future tasks, we promote sparsity on the set of attention vectors $\se{A}^t = \{ \ve{a}^t_1,\dots \ve{a}^t_{L-1} \}$. To do so, we add a regularization term to the loss function $\ab{L}$ that takes into account the set of cumulative attention vectors up to task $t-1$, $\se{A}^{<t} = \{ \ve{a}^{<t}_1,\dots \ve{a}^{<t}_{L-1} \}$:
\begin{equation}
	\ab{L}^\prime\left(\ve{y},\hat{\ve{y}},\se{A}^t,\se{A}^{<t}\right) = \ab{L}\left(\ve{y},\hat{\ve{y}}\right)  + c R\left( \se{A}^t , \se{A}^{<t} \right) ,
    \label{eq:loss}
\end{equation}
where $c$ is the regularization constant, 
\begin{equation}
	R\left( \se{A}^t,\se{A}^{<t} \right) = \frac{ \sum_{l=1}^{L-1} \sum_{i=1}^{N_l} a^t_{l,i} \left( 1-a^{<t}_{l,i} \right) }{ \sum_{l=1}^{L-1} \sum_{i=1}^{N_l} 1-a^{<t}_{l,i} } 
    \label{eq:reg}
\end{equation}
is the regularization term, and $N_l$ corresponds to the number of units in layer $l$. Notice that Eq.~\ref{eq:reg} corresponds to a weighted and normalized L1~regularization over $\se{A}^t$. Cumulative attentions over the past tasks $\se{A}^{<t}$ define a weight for the current task, such that if $a^{<t}_{l,i} \to 1$ then $a^t_{l,i}$ receives a weight close to 0 and vice versa. This excludes the units that were attended in previous tasks from regularization, unconstraining their reuse in the current task. 
The hyperparameter $c\geq 0$ controls the capacity spent on each task (Eq.~\ref{eq:loss}). In a sense, it can be thought of as a compressibility constant, affecting the compactness of the learned models: the higher the $c$, the lower the number of active attention values $a^t_{l,i}$ and the more sparse the resulting network is. We set $c$ globally for all tasks and let HAT adapt to the best compression for each individual task. 

The use of L1~regularization to promote network sparsity in the context of catastrophic forgetting has also been considered by~\citet{Yoon18ICLR} with dynamically expandable networks (DEN), which were introduced while developing HAT. In DEN, plain L1~regularization is combined with a considerable set of heuristics such as L2-transfer, thresholding, and a measure of ``semantic drift'', and is applied to all network weights in the so-called ``selective retraining'' phase. In HAT, we use an attention-weighted L1~regularization over attention values, which is an independent part of the single training phase of the approach. Instead of considering network weights, HAT focuses on unit attention.


\section{Related Work}
\label{sec:SOTA}

We compare the proposed approach with the conceptually closest works, some of which appeared concurrently to the development of HAT. A more general overview of related work has been done in Sec.~\ref{sec:Intro}. A qualitative comparison with three of the most related strategies has been done along Sec.~\ref{sec:Method}. A quantitative comparison with these and other approaches is done in Sec.~\ref{sec:Exper} and \apx{apx:Res}. 

Both elastic weight consolidation~\citep[EWC;][]{Kirkpatrick17PNAS} and synaptic intelligence~\citep[SI;][]{Zenke17ICML} approaches add a `soft' structural regularization term to the loss function in order to discourage changes to weights that are important for previous tasks. HAT uses a `hard' structural regularization, and does it both at the loss function and gradient magnitudes explicitly. EWC measures weights' importance after network training, while SI and HAT compute weights' importance concurrently to network training. EWC and SI use specific formulation while HAT learns attention masks. Incremental moment matching~\citep[IMM;][]{Lee17NIPS} is an evolution of EWC, performing a separate model-merging step after learning a new task.

Progressive neural networks~\citep[PNNs;][]{Rusu16ARXIV} distribute the network weights in a column-wise fashion, pre-assigning a column width per task. They employ so-called adapters to reuse knowledge from previous columns/tasks, leading to a progressive increase of the number of weights assigned to future tasks. Instead of blindly pre-assigning column widths, HAT learns such `widths' per layer, together with the network weights, and adapts them to the difficulty of the current task.
PathNet~\cite{Fernando17ARXIV} also pre-assigns some amount of network capacity per task but, in contrast to PNNs, avoids network columns and adapters. It uses an evolutionary approach to learn paths between a constant number of so-called modules (layer subsets) that interconnect between themselves. HAT does not maintain a population of solutions, entirely trains with backpropagation and SGD, and does not rely on a constant set of modules. 

Together with PNNs and PathNet, PackNet~\cite{Mallya17ARXIV} also employs a binary mask to constrain the network. However, such constrain is not based on columns nor layer modules, but on network weights. Therefore, it allows for a potentially better use of the network's capacity. PackNet is based on heuristic weight pruning, with pre-assigned pruning ratios. HAT also focuses on network weights, but uses unit-based masks to constrain those, which also results in a lightweight structure. It avoids any absolute or pre-assigned pruning ratio, although it uses the compressibility parameter $c$ to influence the compactness of the learned models. Another difference between HAT and the previous three approaches is that it does not use purely binary masks. Instead, the stability parameter $s_{\max}$ controls the degree of binarization.
Dynamically expandable networks~\citep[DEN;][]{Yoon18ICLR} also assign network capacity depending on the task at hand. However, they do so in a separate stage called ``selective retraining''. A complex mixture of heuristics and hyperparameters is used to identify ``drifting'' units, which are duplicated and retrained in another stage. L1~regularization and L2-transfer are employed to condition learning, together with the corresponding regularization constants and an additional set of thresholds. HAT strives for simplicity, restricting the number of hyperparameters to two that have a straightforward conceptual interpretation. Instead of plain L1~regularization over network weights, HAT employs an attention-weighted L1~regularization over attention masks. Attention masks are a lightweight structure that can be plugged in without the need of introducing important changes to a pre-existing network.


\section{Experiments}
\label{sec:Exper}

\textbf{Setups ---} Common setups to evaluate catastrophic forgetting in a classification context are based on permutations of the MNIST data~\cite{Srivastava13NIPS}, label splits of the MNIST data~\cite{Lee17NIPS}, incrementally learning classes of the CIFAR data sets~\cite{LopezPaz17NIPS}, or two-task transfer learning setups where accuracy is measured on both source and target tasks~\cite{Li17PAMI}. However, there are some limitations with these setups. Firstly, performing permutations of the MNIST data has been suggested to favor certain approaches, yielding misleading results\footnote{Essentially, the MNIST data contains many values close to 0 that allow for an easier identification of the important units or weights which, if permuted, can then be easily frozen without overlapping with the ones of the other tasks~\citep[see][]{Lee17NIPS}.} in the context of catastrophic forgetting~\cite{Lee17NIPS}. Secondly, using only the MNIST data may not be very representative of modern computer vision tasks, nor particularly challenging~\cite{Xiao17ARXIV}. Thirdly, incrementally adding classes or groups of classes implies the assumption that all data comes from the same joint distribution, which is unrealistic for a real-world setting. Finally, evaluating catastrophic forgetting with only two tasks biases the conclusions towards transfer learning setups, and prevents the analysis of truly sequential learning with more than two tasks. In this paper, we consider the aforementioned MNIST and CIFAR setups (Sec.~\ref{sec:Exper_Addition}). Nonetheless, we primarily evaluate on a sequence of multiple tasks formed by different classification data sets (Sec.~\ref{sec:Exper_Mixture}). 

To obtain a generic estimate, we weigh a number of tasks and uniformly randomize their order. After training task $t$, we compute the accuracies on all testing sets of tasks $\tau\leq t$. We repeat 10~times this sequential train/test procedure with 10~different seed numbers, which are also used in the rest of randomizations and initializations (see below). To compare between different task accuracies, and in order to obtain a general measurement of the amount of forgetting, we introduce the forgetting ratio
\begin{equation}
	\rho^{\tau \leq t} = \frac{ A^{\tau\leq t} - A^\tau_{\text{R}} }{ A^{\tau\leq t}_{\text{J}} - A^\tau_{\text{R}} } -1,
    \label{eq:accuracy}
\end{equation}
where $A^{\tau\leq t}$ is the accuracy measured on task $\tau$ after sequentially learning task $t$, $A^\tau_{\text{R}}$ is the accuracy of a random stratified classifier using the class information of task $\tau$, and $A^{\tau\leq t}_{\text{J}}$ is the accuracy measured on task $\tau$ after jointly learning $t$ tasks in a multitask fashion. Note that $\rho\approx -1$ and $\rho\approx 0$ correspond to performances close to the ones of the random and multitask classifiers, respectively. To report a single number after learning $t$ tasks, we take the average
\begin{equation*}
	\rho^{\leq t} = \frac{1}{t} \sum_{\tau=1}^t \rho^{\tau \leq t} .
\end{equation*}

\textbf{Data ---} We consider 8~common image classification data sets and adapt them, if necessary, to an input size of $32\times 32\times 3$ pixels. The number of classes goes from 10 to~100, training set sizes from 16,853 to~73,257, and test set sizes from 1,873 to~26,032. For each task, we randomly split 15\% of the training set and keep it for validation purposes. The considered data sets are: CIFAR10 and CIFAR100~\cite{krizhevsky2009learning}, FaceScrub~\cite{ng2014data}, FashionMNIST~\cite{Xiao17ARXIV}, NotMNIST~\cite{Bulatov11TR}, MNIST~\cite{LeCun98PIEEE}, SVHN~\cite{Netzer2011}, and TrafficSigns~\cite{Stallkamp2011IJCNN}. For further details on data we refer to \apx{apx:Setup_Data}. 

\textbf{Baselines ---} We consider 2~reference approaches plus 9~recent and competitive ones: standard SGD with dropout~\cite{Goodfellow14ICLR}, SGD freezing all layers except the last one (SGD-F), EWC, IMM (Mean and Mode variants), learning without forgetting~\citep[LWF;][]{Li17PAMI}, less-forgetting learning~\citep[LFL;][]{Jung16ARXIV}, PathNet, and PNNs. To find the best hyperparameter combination for each approach, we perform a grid search using a task sequence determined by a single seed. To compute the forgetting ratio $\rho$ (Eq.~\ref{eq:accuracy}), we also run the aforementioned random and multitask classifiers.

\textbf{Network ---} Unless stated otherwise, we employ an AlexNet-like architecture~\cite{krizhevsky2012imagenet} with 3~convolutional layers of 64, 128, and 256 filters with $4\times 4$, $3\times 3$, and $2\times 2$ kernel sizes, respectively, plus two fully-connected layers of 2048 units each. We use rectified linear units as activations, and $2\times 2$ max-pooling after the convolutional layers. We also use a dropout of 0.2 for the first two layers and of 0.5 for the rest. A fully-connected layer with a softmax output is used as a final layer, together with categorical cross entropy loss. All layers are randomly initialized with Xavier uniform initialization~\cite{glorot2010understanding} except the embedding layers, for which we use a Gaussian distribution $\ab{N}(0,1)$. Unless stated otherwise, our code uses PyTorch's defaults for version 0.2.0~\cite{Paszke17NIPSW}. We adapt the same base architecture to all baseline approaches and match their number of parameters to 7.1\,M. 

\textbf{Training ---} We train all models with backpropagation and plain SGD, using a learning rate of 0.05, and decaying it by a factor of 3 if there is no improvement in the validation loss for 5~consecutive epochs. We stop training when we reach a learning rate lower than $10^{-4}$ or we have iterated over 200~epochs (we made sure that all considered approaches reached a stable solution before 200~epochs). Batch size is set to 64. All methods use the same task sequence, data split, batch shuffle, and weight initialization for a given seed. 

\subsection{Results}
\label{sec:Exper_Mixture}

We first look at the average forgetting ratio $\rho^{\leq t}$ after learning task $t$ (Fig.~\ref{fig:mixture}). A first thing to note is that not all the considered baselines perform better than the SGD references. That is the case of LWF and LFL. For LWF, we observe it is still competitive in the two-task setup for which it was designed, $t=2$. However, its performance rapidly degrades for $t>2$, indicating that the approach has difficulties in extending beyond a transfer learning setup. We find LFL extremely sensitive to the configuration of its hyperparameter, to the point that what is a good value for one seed, turns out to be a bad choice for another seed. Hence the poor average performance for 10~seeds. The highest standard deviations are obtained by LFL and PathNet (Table~\ref{tab:mixture0}), which suggests a high sensitivity with respect to hyperparameters, initializations, or data sets. Another thing to note is that the IMM approaches only perform similarly or slightly better than the SGD-F reference. We believe this is due to both the different nature of the tasks' data and the consideration of more than two tasks, which complicates the choice of the mixing hyperparameter.

\begin{figure}[t]
	\begin{center}
	\includegraphics{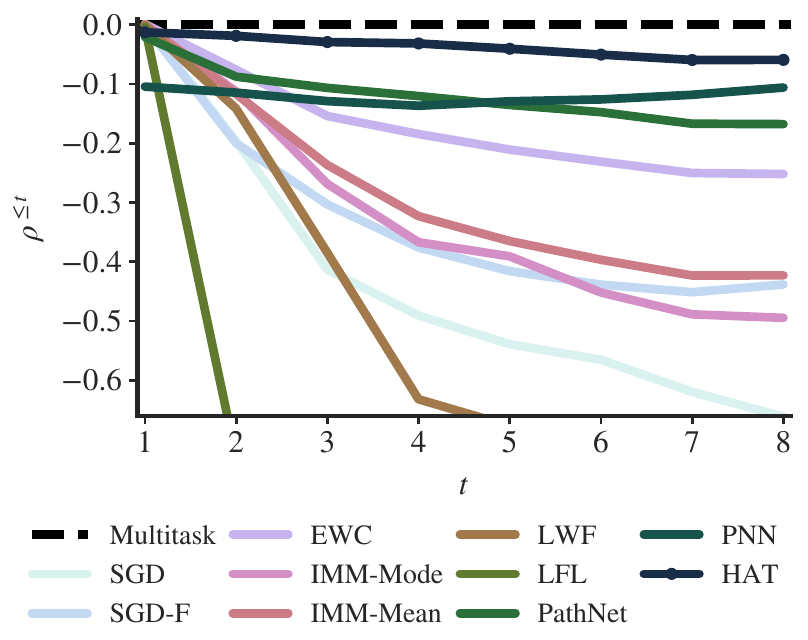}
    \vskip -0.15in
	\caption{Average forgetting ratio $\rho^{\leq t}$ for the considered approaches (10~runs).}
	\label{fig:mixture}
	\end{center}
	\vskip -0.1in
\end{figure} 

\begin{table}[t]
    \caption{Average forgetting ratio after the second ($\rho^{\leq 2}$) and the last ($\rho^{\leq 8}$) task for the considered approaches (10~runs, standard deviation into parenthesis).}
    \label{tab:mixture0}
    \begin{center}
    \begin{small}
    \begin{sc}
    \begin{tabular}{lcc}
    \toprule
    Approach & $\rho^{\leq 2}$ & $\rho^{\leq 8}$ \\
    \midrule
LFL          & -0.73 (0.29) & -0.92 (0.08) \\
LWF          & -0.14 (0.13) & -0.80 (0.06) \\
SGD          & -0.20 (0.08) & -0.66 (0.03) \\
IMM-Mode     & -0.11 (0.08) & -0.49 (0.05) \\
SGD-F        & -0.20 (0.15) & -0.44 (0.06) \\
IMM-Mean     & -0.12 (0.10) & -0.42 (0.04) \\
EWC          & -0.08 (0.06) & -0.25 (0.03) \\
PathNet      & -0.09 (0.16) & -0.17 (0.23) \\
PNN          & -0.11 (0.10) & -0.11 (0.01) \\
\textbf{HAT}         & \textbf{-0.02 (0.03)} & \textbf{-0.06 (0.01)} \\
	\bottomrule
    \end{tabular}
    \end{sc}
    \end{small}
    \end{center}
    \vskip -0.1in
\end{table}

The best performing baselines are EWC, PathNet, and PNN. PathNet and PNN present contrasting behaviors. Both, by construction, never forget; therefore, the important difference is in their learning capability. PathNet starts by correctly learning the first task and progressively exhibits difficulties to do so for $t\geq 2$. Contrastingly, PNNs exhibits difficulty in the first tasks and becomes better as $t$ increases. These contrasting behaviors are due to the way the two approaches allocate the network capacity. As mentioned, they cannot do it dynamically, and therefore need to pre-assign a number of network weights per task. When having more tasks but the same network capacity, this pre-assignment increasingly harms the performance of these baselines, lowering the corresponding curves in Fig.~\ref{fig:mixture}.

We now move to the HAT results. First of all, we observe that HAT consistently performs better than all considered baselines for all $t\geq 2$ (Fig.~\ref{fig:mixture}). For the case of $t=2$, it obtains an average forgetting ratio $\rho^{\leq 2}=-0.02$, while the best baseline is EWC with $\rho^{\leq 2}=-0.08$ (Table~\ref{tab:mixture0}). For the case of $t=8$, HAT obtains $\rho^{\leq 8}=-0.06$, while the best baseline is PNN with $\rho^{\leq 8}=-0.11$. This implies a reduction in forgetting of 75\% for $t=2$ and 45\% for $t=8$. Notice that the standard deviation of HAT is lower than the ones obtained by the big majority of the baselines (Table~\ref{tab:mixture0}). This denotes a certain stability of HAT with respect to different task sequences, data sets, data splits, and network initializations. 

Given the slightly increasing tendency of PNN with $t$ (Fig.~\ref{fig:mixture}), one could speculate that PNN would score above HAT for $t>8$. However, our empirical analyzes suggest that that is not the case (presumably due to the capacity pre-assignment and parameter increase problems underlined in Sec.~\ref{sec:SOTA} and above). In particular, we observe a gradual lowering of PathNet and PNN curves with increasing sequences from $t=2$~to~8. In addition, we observe PathNet and PNN obtaining worse performances than EWC in the case of $t=10$ for the incremental class setup (see below and \apx{apx:Res_IncCIFAR}). In general, none of the baseline methods consistently outperforms the rest across setups and for all $t$, a situation that we do observe with HAT. 

\subsection{Additional Results}
\label{sec:Exper_Addition}

To broaden the strength of our results, we additionally experiment with three common alternative setups. First, we consider an incremental class learning scenario, similar to~\citet{LopezPaz17NIPS}, using class subsets of both CIFAR10 and CIFAR100 data. In this setup, the best baseline after $t\geq 3$ is EWC, with $\rho^{\leq 10}=-0.18$. HAT scores $\rho^{\leq 10}=-0.09$ (55\% forgetting reduction). Next, we consider the permuted MNIST sequence of tasks~\cite{Srivastava13NIPS}. In this setup, the best result we could find in the literature was from SI, with $A^{\leq 10}=97.1\%$. HAT scores $A^{\leq 10}=98.6\%$ (52\% error rate reduction). Finally, we also consider the split MNIST task of~\citet{Lee17NIPS}. In this setup, the best result from the literature corresponds to the conceptor-aided backpropagation approach~\cite{He18ICLR}, with $A^{\leq 2}=94.9\%$. HAT scores $A^{\leq 2}=99.0\%$ (80\% error rate reduction). The detail for all these setups and results can be found in \apx{apx:Res}.

\subsection{Hyperparameters}
\label{sec:Exper_Param}

In any machine learning algorithm, it is important to assess the sensitivity with respect to the hyperparameters. HAT has two: the stability parameter $s_{\max}$ and the compressibility parameter $c$ (Secs.~\ref{sec:Method_AttTrain} and~\ref{sec:Method_Compress}). A low $s_{\max}$ provides plasticity to the units and capacity of adaptation, but the network may easily forget what it learned. A high $s_{\max}$ prevents forgetting, but the network may have difficulties in adapting to new tasks. A low $c$ allows to use almost all of the network's capacity for a given task, potentially spending too much in the current task. A high $c$ forces it to learn a very compact model, at the expense of not reaching the accuracy that the original network could have reached. We empirically found good operation ranges $s_{\max} \in [25,800]$ and $c \in [0.1,2.5]$. As we can see, any variation within these ranges results in reasonable performance (Fig.~\ref{fig:params}). Unless stated otherwise, we use $s_{\max}=400$ and $c=0.75$.

\begin{figure}[t]
	\begin{center}
	\includegraphics{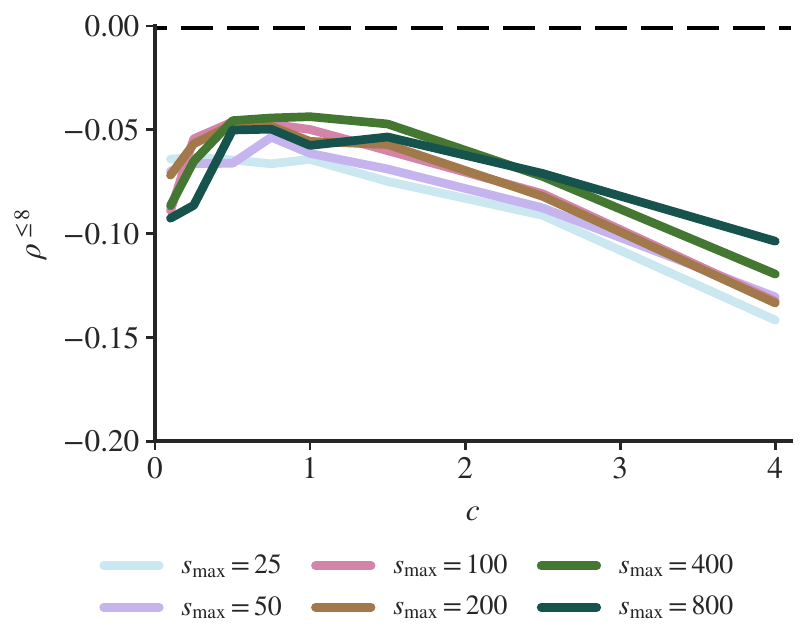}
    \vskip -0.15in
	\caption{Effect of hyperparameters $s_{\max}$ and $c$ on average forgetting ratio $\rho^{\leq8}$. Results for seed 0.}
	\label{fig:params}
	\end{center}
	\vskip -0.1in
\end{figure} 

\subsection{Monitoring and Network Pruning}
\label{sec:Exper_Monitor}

It is interesting to note that the hard attention mechanism introduced in Sec.~\ref{sec:Method} offers a number of possibilities to monitor the behavior of our models. For instance, by computing the conditioning mask in Eq.~\ref{eq:gradmask} from the hard attention vectors $\ve{a}^{\leq t}_l$, we can assess which weights obtain a high attention value, binarize it, and compute an estimate of the instantaneous network capacity usage (Fig.~\ref{fig:capacity}). We may also inform ourselves of the amount of active weights per layer and task (\apx{apx:layeruse}). Another facet we can monitor is the weight reuse across tasks. By a similar procedure, comparing the conditioning masks between tasks $t_i$ and $t_j$, $j>i$, we can asses the percentage of weights of task $t_i$ that are later reused in task $t_j$ (Fig.~\ref{fig:neuron_overlap}).

\begin{figure}[t]
	\begin{center}
	\includegraphics{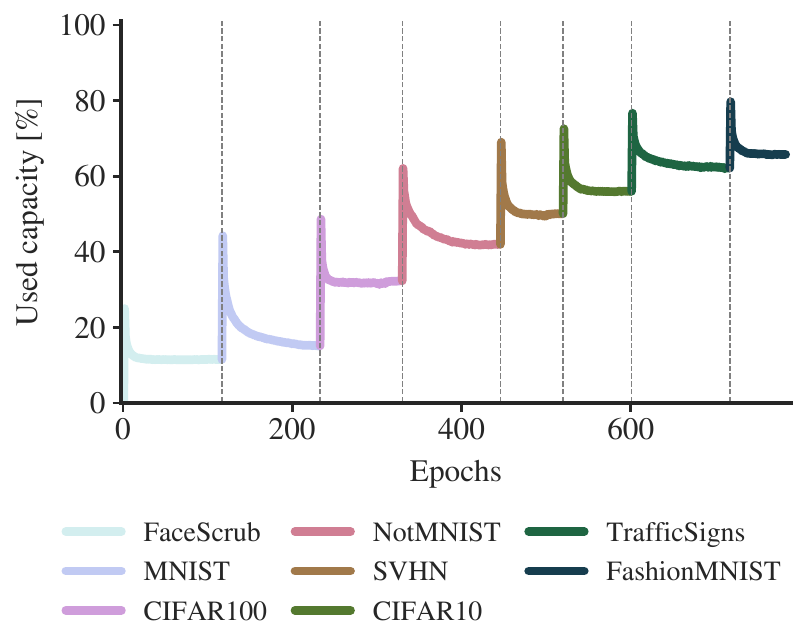}
    \vskip -0.15in
	\caption{Network capacity usage with sequential task learning (seed 0). Dashed vertical lines correspond to a task switch.}
	\label{fig:capacity}
	\end{center}
	\vskip -0.1in
\end{figure} 

\begin{figure}[t]
	\vskip 0.2in
	\begin{center}
	\includegraphics{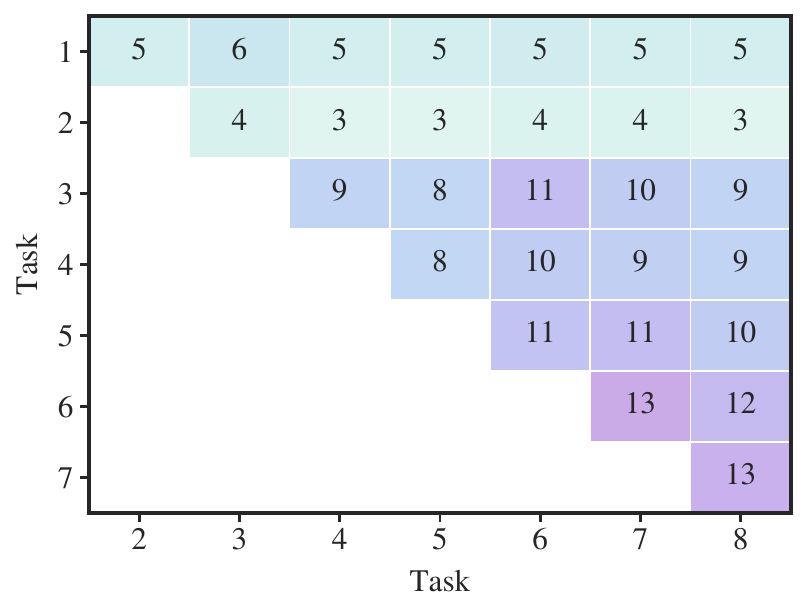}
    \vskip -0.1in
	\caption{Percentage of weight reuse across tasks. Seed 0 sequence: (1) FaceScrub, (2) MNIST, (3) CIFAR100, (4) NotMNIST, (5) SVHN, (6) CIFAR10, (7) TrafficSigns, and (8) FashionMNIST.}
	\label{fig:neuron_overlap}
	\end{center}
	\vskip -0.1in
\end{figure} 

Another by-product of hard attention masks is that we can use them to assess which of the network's weights are important, and then prune the most irrelevant ones~\cite{LeCun90NIPS}. This way, we can compress the network for further deployment in low-resource devices or time-constrained environments~\citep[cf.][]{Han16ICLR}. If we want to focus on such compression task, we can set $c$ to a higher value than the one used for catastrophic forgetting and start with a positive random initialization of the embeddings $\ve{e}_l$. The former will promote more compression while the latter will ensure we start learning the model by putting attention to all weights in the first epochs (full capacity). We empirically found that using $c=1.5$ and $\ab{U}(0,2)$ yields a reasonable trade-off between accuracy and compression for a single task (Fig.~\ref{fig:compression}). With that, we can compress the network to sizes between 1 and 21\% of its original size, depending on the task (\apx{apx:Raw_Compress}). Comparing these numbers with the compression rates used by PackNet (25 or 50\%), we see that HAT generally uses a much more compact model. Comparing with DEN on the specific MNIST and CIFAR100 tasks (18 and 52\%), we observe that HAT compresses to 1 and 21\%, respectively. Interestingly, and in contrast to these and the majority of network pruning approaches, HAT learns to prune network weights through backpropagation and SGD, and at the same time as the network weights themselves.

\begin{figure}[t]
	\begin{center}
	\includegraphics{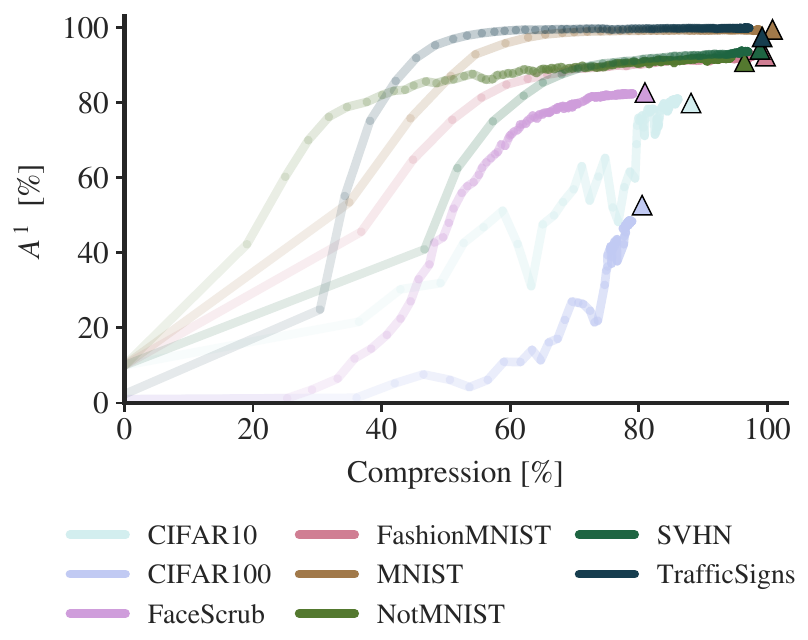}
    \vskip -0.15in
	\caption{Validation accuracy $A^1$ as a function of compression percentage. Every dot corresponds to an epoch and triangles match the accuracy of the SGD approach (no compression).}
	\label{fig:compression}
	\end{center}
	\vskip -0.1in
\end{figure} 


\section{Conclusion}
\label{sec:Conclusion}

We introduce HAT, a hard attention mechanism that, by focusing on a task embedding, is able to protect the information of previous tasks while learning new tasks. This hard attention mechanism is lightweight, in the sense that it adds a small fraction of weights to the base network, and is trained together with the main model, with negligible overhead using backpropagation and vanilla SGD. We demonstrate the effectiveness of the approach to control catastrophic forgetting in the image classification context by running a series of experiments with multiple data sets and state-of-the-art approaches. HAT has only two hyperparameters, which intuitively refer to the stability and compactness of the learned knowledge, and whose tuning we demonstrate is not crucial for obtaining good performance. In addition, HAT offers the possibility to monitor the used network capacity across tasks and layers, the unit reuse across tasks, and the compressibility of a model trained for a given task. We hope that our approach may be also useful in online learning or network compression contexts, and that the hard attention mechanism presented here may also find some applicability beyond the catastrophic forgetting problem.


 


\balance
\bibliography{biblio}
\bibliographystyle{icml2018}

	}{
    	\onecolumn
\appendix

\iftoggle{twodoc}{
	\icmltitle{\titletext}
    \vspace*{0.05cm}
    \begin{center}
	\textbf{\Large SUPPLEMENTARY MATERIALS}
    \end{center}
    \vspace*{0.3cm}
}{
	\vspace*{-0.7cm}
	\section*{APPENDIX}
    \vspace*{0.2cm}
}



\section{Data}
\label{apx:Setup_Data}

The data sets used in our experiments are summarized in Table~\ref{tab:data sets}. The MNIST data set~\cite{LeCun98PIEEE} comprises $28\times 28$ monochromatic images of handwritten digits. 
Fashion-MNIST~\cite{Xiao17ARXIV} comprises gray-scale images of the same size from Zalando's articles\footnote{\url{https://github.com/zalandoresearch/fashion-mnist}}.
The German traffic sign data set~\citep[TrafficSigns;][]{Stallkamp2011IJCNN} contains traffic sign images. We used the version of the data set from the Udacity self-driving car github repository\footnote{\url{https://github.com/georgesung/traffic_sign_classification_german}}. 
The NotMNIST data set~\cite{Bulatov11TR} comprises glyphs extracted from publicly available fonts, making a similar data set to MNIST; we just need to resize the images\footnote{Code and processed data available on github: \url{https://github.com/nkundiushuti/notmnist_convert}
}.
The SVHN data set~\cite{Netzer2011} comprises digits cropped from house numbers in Google Street View images.
The FaceScrub data set~\cite{ng2014data} is widely used in face recognition tasks~\cite{kemelmacher2016megaface}. Because some of the images listed in the original data set were not hosted anymore on the corresponding Internet domains, we use a version of the data set stored on the MegaFace challenge website\footnote{\url{http://megaface.cs.washington.edu/participate/challenge.html}}~\cite{kemelmacher2016megaface}, from which we select the first 100 people with the most appearances\footnote{Code and processed data available on github: \url{https://github.com/nkundiushuti/facescrub_subset}
}. The CIFAR10 and CIFAR100 data sets contain $32\times 32$ color images~\cite{krizhevsky2009learning}.

To match the image input shape required in our experiments, some of the images in the corresponding data sets need to be resized (FaceScrub, TrafficSigns, and NotMNIST) or padded with zeros (MNIST and FashionMNIST). In addition, for the data sets comprising monochromatic images, we replicate the image across all RGB channels. Note that we do not perform any sort of data augmentation; we just adapt the inputs. We provide the necessary code to perform such adaptations in the links listed above.

\begin{table}[h!t]
    \caption{Data sets used in the study: name, reference, number of classes, and number of train and test instances.}
    \label{tab:data sets}
    \vskip 0.15in
    \begin{center}
    \begin{small}
    \begin{sc}
    \begin{tabular}{lcrr}
    \toprule
    Data set & Classes & Train & Test \\
    \midrule
    CIFAR10~\cite{krizhevsky2009learning}	& 10 & 50,000 & 10,000 \\
    CIFAR100~\cite{krizhevsky2009learning}	& 100 & 50,000 & 10,000 \\
    FaceScrub~\cite{ng2014data}				& 100 & 20,600 & 2,289 \\
    FashionMNIST~\cite{Xiao17ARXIV}			& 10 & 60,000 & 10,000 \\
    NotMNIST~\cite{Bulatov11TR}				& 10 & 16,853 & 1,873 \\
	MNIST~\cite{LeCun98PIEEE}				& 10 & 60,000 & 10,000 \\
    SVHN~\cite{Netzer2011}					& 100 & 73,257 & 26,032 \\
    TrafficSigns~\cite{Stallkamp2011IJCNN}	& 43 & 39,209 & 12,630 \\
    \bottomrule
    \end{tabular}
    \end{sc}
    \end{small}
    \end{center}
    \vskip -0.1in
\end{table}

\section{Raw Results}
\label{apx:Raw}

\subsection{Task Mixture}
\label{apx:Raw_Mixture}

We report all forgetting ratios $\rho^{\leq t}$ for $t=1$ to 8 in Table~\ref{tab:mixture1}. A total of 10~runs with 10~different seeds are performed and the averages and standard deviations are taken.

\begin{table}[h!t]
    \caption{Average forgetting ratio $\rho^{\leq t}$ for the considered approaches (10~runs, standard deviation into parenthesis).}
    \label{tab:mixture1}
    \vskip 0.15in
    \begin{center}
    \begin{small}
    \begin{sc}
    \resizebox{\textwidth}{!}{
    \begin{tabular}{lcccccccc}
    \toprule
    Approach & $\rho^{\leq 1}$ & $\rho^{\leq 2}$ & $\rho^{\leq 3}$ & $\rho^{\leq 4}$ & $\rho^{\leq 5}$ & $\rho^{\leq 6}$ & $\rho^{\leq 7}$ & $\rho^{\leq 8}$ \\
    \midrule
LFL          & -0.00 (0.01) & -0.73 (0.29) & -0.88 (0.18) & -0.89 (0.13) & -0.91 (0.11) & -0.90 (0.09) & -0.92 (0.08) & -0.92 (0.08) \\
LWF          & -0.00 (0.01) & -0.14 (0.13) & -0.38 (0.17) & -0.63 (0.11) & -0.68 (0.08) & -0.70 (0.03) & -0.76 (0.06) & -0.80 (0.06) \\
SGD          &  -0.00 (0.00) & -0.20 (0.08) & -0.41 (0.09) & -0.49 (0.07) & -0.54 (0.07) & -0.57 (0.06) & -0.62 (0.06) & -0.66 (0.03) \\
IMM-Mode     & -0.00 (0.01) & -0.11 (0.08) & -0.27 (0.12) & -0.37 (0.10) & -0.39 (0.07) & -0.45 (0.05) & -0.49 (0.06) & -0.49 (0.05) \\
SGD-F        & -0.00 (0.00) & -0.20 (0.15) & -0.30 (0.15) & -0.38 (0.11) & -0.42 (0.09) & -0.44 (0.08) & -0.45 (0.07) & -0.44 (0.06) \\
IMM-Mean     &  -0.00 (0.00) & -0.12 (0.10) & -0.24 (0.11) & -0.32 (0.06) & -0.37 (0.06) & -0.40 (0.06) & -0.42 (0.07) & -0.42 (0.04) \\
EWC          &  -0.00 (0.00) & -0.08 (0.06) & -0.15 (0.11) & -0.18 (0.07) & -0.21 (0.07) & -0.23 (0.04) & -0.25 (0.05) & -0.25 (0.03) \\
PathNet      & -0.02 (0.03) & -0.09 (0.16) & -0.11 (0.19) & -0.12 (0.21) & -0.14 (0.22) & -0.15 (0.23) & -0.17 (0.23) & -0.17 (0.23) \\
PNN          & -0.10 (0.12) & -0.11 (0.10) & -0.13 (0.09) & -0.14 (0.04) & -0.13 (0.03) & -0.13 (0.02) & -0.12 (0.01) & -0.11 (0.01) \\
\textbf{HAT}          & \textbf{-0.01 (0.02)} & \textbf{-0.02 (0.03)} & \textbf{-0.03 (0.03)} & \textbf{-0.03 (0.02)} & \textbf{-0.04 (0.02)} & \textbf{-0.05 (0.02)} & \textbf{-0.06 (0.02)} & \textbf{-0.06 (0.01)} \\
	\bottomrule
    \end{tabular}
    }
    \end{sc}
    \end{small}
    \end{center}
    \vskip -0.1in
\end{table}




\subsection{Layer Use}
\label{apx:layeruse}

In Fig.~\ref{fig:layeruse} we show an example of layer capacity monitoring as the sequence of tasks evolves. As mentioned in the main paper, we can compute a percent of active weights for a given layer and task.

\begin{figure}[h!t]
	\begin{center}
    \includegraphics[width=0.55\linewidth]{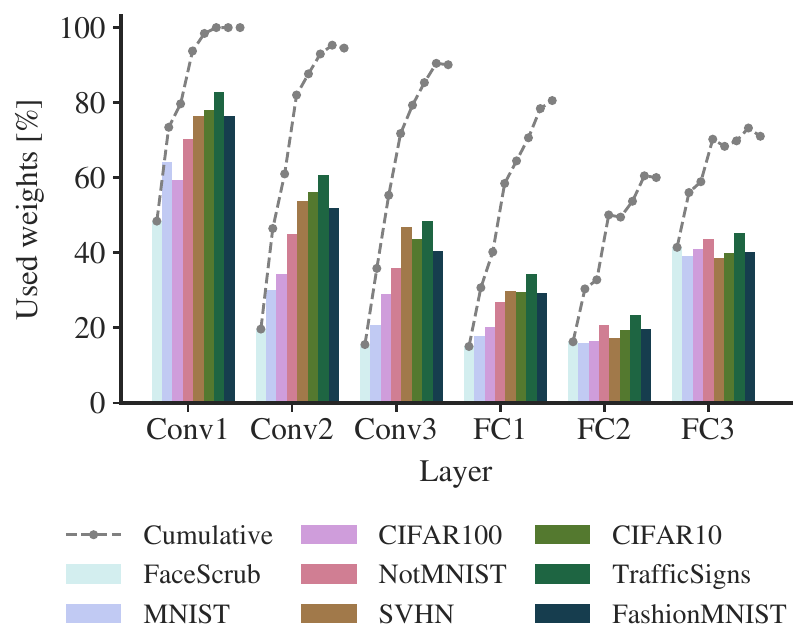}
    \vskip -0.15in
	\caption{Layer-wise weight usage with sequential task learning, including (lines) and excluding (bars) the cumulative attention of past tasks. Task sequence corresponds to seed~0.}
	\label{fig:layeruse}
	\end{center}
	\vskip -0.1in
\end{figure} 

\subsection{Network Compression}
\label{apx:Raw_Compress}

The final results of the network compression experiment reported in the main paper (after reaching convergence) are available in Table~\ref{tab:Compress}. We run HAT on isolated tasks with $c=1.5$ and uniform embedding initialization $\ab{U}(0,2)$.

\begin{table}[h!t]
    \caption{Results for the compression experiment reported in the main paper: test accuracy $A^1$ with SGD, test accuracy $A^1$ after compressing with HAT, and percentage of network weights used after compression.}
    \label{tab:Compress}
    \vskip 0.15in
    \begin{center}
    \begin{small}
    \begin{sc}
    \begin{tabular}{lccr}
    \toprule
    Data set 		& Raw $A^1$ & Compressed $A^1$ 	& Size \\
    \midrule
	CIFAR10			& 79.9\%	& 80.8\%			& 13.9\% 	\\
    CIFAR100		& 52.7\%	& 49.1\%			& 21.4\% 	\\
    FaceScrub		& 82.7\%	& 82.3\%			& 21.0\% 	\\
    FashionMNIST	& 92.4\%	& 91.9\%			& 2.3\%		\\
    MNIST			& 99.5\%	& 99.4\% 			& 1.2\%		\\
    NotMNIST		& 90.9\%	& 91.5\%			& 5.7\%		\\
    SVHN			& 94.2\%	& 93.8\%			& 3.1\%		\\
    TrafficSigns	& 97.5\%	& 98.1\%			& 2.9\%		\\
	\bottomrule
    \end{tabular}
    \end{sc}
    \end{small}
    \end{center}
    \vskip -0.1in
\end{table}

\subsection{Training Time}
\label{apx:Raw_Time}

To have an idea of the training time for each of the considered approaches, we report some reference values in Table~\ref{tab:Time}. We see that HAT is also quite competitive in this aspect.

\begin{table}[h!t]
    \caption{Wall-clock training time measured on a single NVIDIA Pascal Titan X GPU: total (after learning the 8~tasks), per epoch, and per batch (batches of 64). Batch processing time is measured for a forward pass (Batch-F), and for both a forward and a backward pass (Batch-FB).}
    \label{tab:Time}
    \vskip 0.15in
    \begin{center}
    \begin{small}
    \begin{sc}
    \begin{tabular}{p{2.4cm}rrrr}
    \toprule
    Approach 	& \multicolumn{4}{c}{Training time} \\
    			& Total [h] & Epoch [s] & Batch-F [ms] & Batch-FB [ms] \\
    \midrule
	PNN				& 6.0 & 4.1 & 10.2 & 27.5 \\
	PathNet			& 4.5 & 3.6 & 10.6 & 23.9 \\
	EWC				& 3.9 & 3.1 & 7.9 & 19.7 \\
	Multitask		& 3.4 & 94.8 & 3.1 & 15.7 \\
	IMM-Mean		& 3.2 & 2.6 & 6.9 & 17.1 \\
	IMM-Mode		& 3.1 & 2.5 & 6.7 & 16.0 \\
	LWF				& 2.2 & 2.2 & 5.7 & 14.2 \\
	\textbf{HAT} 	& \textbf{2.2} & \textbf{1.6} & \textbf{4.0} & \textbf{11.7} \\
	SGD				& 1.4 & 0.9 & 2.5 & 6.6 \\
	LFL				& 1.3 & 0.9 & 4.4 & 9.2 \\
	SGD-F			& 0.5 & 0.9 & 2.5 & 6.8 \\
	\bottomrule
    \end{tabular}
    \end{sc}
    \end{small}
    \end{center}
    \vskip -0.1in
\end{table}


\section{Additional Results}
\label{apx:Res}

\subsection{Incremental CIFAR}
\label{apx:Res_IncCIFAR}

As an additional experiment to complement our evaluation, we consider the incremental CIFAR setup, following a similar approach as~\citet{LopezPaz17NIPS}. We divide both CIFAR10 and CIFAR100 data sets into consecutive-class subsets and use them as tasks, presented in random order according to the seed. We take groups of 2~classes for CIFAR10 and 20~classes for CIFAR100, yielding a total of 10~tasks. We decide to take groups of 2 and 20~classes in order to have a similar number of training instances per task. The rest of the procedure is as in the main paper. The most important results are summarized there. The complete numbers are depicted in Fig.~\ref{fig:cifar} and reported in Table~\ref{tab:cifar1}.

\begin{figure}[h!t]
	\vskip 0.2in
	\begin{center}
	\includegraphics[width=0.55\linewidth]{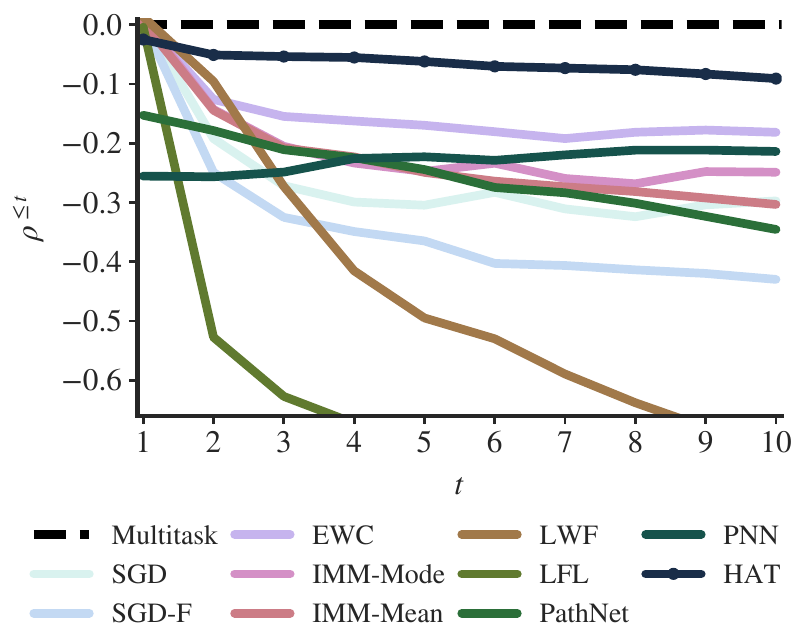}
    \vskip -0.1in
	\caption{Average forgetting ratio $\rho^{\leq t}$ for the incremental CIFAR task (average after 10~runs).}
	\label{fig:cifar}
	\end{center}
	\vskip -0.1in
\end{figure} 

\begin{table}[h!t]
    \caption{Average forgetting ratio $\rho^{\leq t}$ for the incremental CIFAR task (10~runs, standard deviation into parenthesis).}
    \label{tab:cifar1}
    \vskip 0.15in
    \begin{center}
    \begin{small}
    \begin{sc}
    \resizebox{\textwidth}{!}{
    \setlength{\tabcolsep}{2pt}
    \begin{tabular}{lcccccccccc}
    \toprule
    Approach & $\rho^{\leq 1}$ & $\rho^{\leq 2}$ & $\rho^{\leq 3}$ & $\rho^{\leq 4}$ & $\rho^{\leq 5}$ & $\rho^{\leq 6}$ & $\rho^{\leq 7}$ & $\rho^{\leq 8}$ & $\rho^{\leq 9}$ & $\rho^{\leq 10}$ \\
    \midrule
LFL          & -0.00 (0.01) & -0.53 (0.31) & -0.63 (0.25) & -0.67 (0.21) & -0.70 (0.20) & -0.74 (0.17) & -0.77 (0.15) & -0.79 (0.14) & -0.79 (0.14) & -0.78 (0.14) \\
LWF          &  -0.00 (0.02) & -0.10 (0.03) & -0.27 (0.05) & -0.42 (0.05) & -0.50 (0.06) & -0.53 (0.04) & -0.59 (0.06) & -0.64 (0.06) & -0.68 (0.05) & -0.70 (0.05) \\
SGD-F        &  -0.00 (0.01) & -0.25 (0.14) & -0.33 (0.16) & -0.35 (0.18) & -0.37 (0.16) & -0.40 (0.18) & -0.41 (0.18) & -0.41 (0.19) & -0.42 (0.19) & -0.43 (0.20) \\
PathNet      & -0.15 (0.31) & -0.18 (0.20) & -0.21 (0.26) & -0.22 (0.28) & -0.24 (0.29) & -0.27 (0.29) & -0.28 (0.30) & -0.30 (0.30) & -0.32 (0.29) & -0.35 (0.28) \\
SGD          &  -0.00 (0.01) & -0.19 (0.09) & -0.27 (0.09) & -0.30 (0.04) & -0.30 (0.06) & -0.28 (0.04) & -0.31 (0.03) & -0.32 (0.04) & -0.30 (0.05) & -0.30 (0.04) \\
IMM-Mean     &  -0.00 (0.02) & -0.14 (0.08) & -0.21 (0.10) & -0.22 (0.10) & -0.25 (0.10) & -0.26 (0.08) & -0.27 (0.08) & -0.28 (0.08) & -0.29 (0.07) & -0.30 (0.07) \\
IMM-Mode     &  -0.00 (0.01) & -0.14 (0.10) & -0.21 (0.11) & -0.23 (0.06) & -0.25 (0.09) & -0.23 (0.07) & -0.26 (0.05) & -0.27 (0.04) & -0.25 (0.04) & -0.25 (0.04) \\
PNN          & -0.26 (0.16) & -0.26 (0.08) & -0.25 (0.05) & -0.23 (0.04) & -0.22 (0.03) & -0.23 (0.03) & -0.22 (0.03) & -0.21 (0.02) & -0.21 (0.02) & -0.21 (0.02) \\
EWC          &  -0.00 (0.01) & -0.13 (0.09) & -0.15 (0.08) & -0.16 (0.07) & -0.17 (0.06) & -0.18 (0.06) & -0.19 (0.08) & -0.18 (0.07) & -0.18 (0.06) & -0.18 (0.06) \\
\textbf{HAT}          & \textbf{-0.03 (0.04)} & \textbf{-0.05 (0.02)} & \textbf{-0.05 (0.02)} & \textbf{-0.06 (0.01)} & \textbf{-0.06 (0.01)} & \textbf{-0.07 (0.01)} & \textbf{-0.07 (0.01)} & \textbf{-0.08 (0.01)} & \textbf{-0.08 (0.01)} & \textbf{-0.09 (0.01)} \\    
	\bottomrule
    \end{tabular}
    }
    \end{sc}
    \end{small}
    \end{center}
    \vskip -0.1in
\end{table}

\subsection{Permuted MNIST}
\label{apx:Res_PMNIST}

A common experiment is the one proposed by~\citet{Srivastava13NIPS}, and later employed to evaluate catastrophic forgetting by~\citet{Goodfellow14ICLR}. It consists of taking random permutations of the pixels in the MNIST data set as tasks. Typically, the average accuracy after sequentially training on 10~MNIST permutations is reported. 
To match the different number of parameters used in the literature, we consider a small, medium, and a large network based on a two-layer fully-connected architecture as~\citet{Zenke17ICML}, with 100, 500, and 2000 hidden units, respectively. For the large network we set dropout probabilities as~\citet{Kirkpatrick17PNAS}. We use $s_{\max}=200$ and $c=0.5$ for the small network, and $s_{\max}=400$ and $c=0.5$ for the medium and large networks. The results are available in Table~\ref{tab:PMNIST}.

\begin{table}[t]
    \caption{Accuracy on the permuted MNIST task~\cite{Srivastava13NIPS}, taking the average after training 10~tasks. The only exception is the generative replay approach, whose performance was assessed after 5~tasks. Superscripts indicate results reported by (1)~\citet{Nguyen17ARXIV} and (2)~\citet{He18ICLR}. An asterisk after parameter count indicates that the approach presents some additional structure not included in such parameter count (for instance, some memory module or an additional generative network).}
    \label{tab:PMNIST}
    \vskip 0.15in
    \begin{center}
    \begin{small}
    \begin{sc}
    \begin{tabular}{lcc}
    \toprule
    Approach & Parameters & $A^{\leq 10}$ \\
    \midrule
    GEM~\cite{LopezPaz17NIPS}           				& 0.1\,M*    & 82.8\% \\
    SI~\cite{Zenke17ICML}$^1$        					& 0.1\,M~   & 86.0\% \\
    EWC~\cite{Kirkpatrick17PNAS}$^2$       				& 0.1\,M~   & 88.2\%  \\
    MbPA + EWC -- 1000 ex.~\cite{Sprechmann18ICLR}      & Unknown*    & 89.7\%  \\
    VCL~\cite{Nguyen17ARXIV}        					& 0.1\,M*    & 90.0\%  \\
    \textbf{HAT -- Small}		            			& \textbf{0.1\,M}~ & \textbf{91.6\%} \\
    Generative Replay~\cite{Shin17NIPS} 				& Unknown*	& 94.9\% \\
    CAB~\cite{He18ICLR}               					& 0.7\,M~    & 95.2\% \\
    EWC~\cite{Kirkpatrick17PNAS}     					& 5.8\,M~    & 96.9\%  \\
    SI~\cite{Zenke17ICML}        						& 5.8\,M~    & 97.1\% \\
    \textbf{HAT -- Medium}              				& \textbf{0.7\,M}~    & \textbf{97.4\%} \\
    \textbf{HAT -- Large}              					& \textbf{5.8\,M}~    & \textbf{98.6\%} \\
    \bottomrule
    \end{tabular}
    \end{sc}
    \end{small}
    \end{center}
    \vskip -0.1in
\end{table}

\subsection{Split MNIST}
\label{apx:Res_MNIST5}

Another popular experiment is to split the MNIST data set into tasks and report the average accuracy after learning them one after the other. We follow~\citet{Lee17NIPS} by splitting the data set using labels 0--4 and 5--9 as tasks and running the experiment 10~times. We also match the base network architecture to the one used by~\citet{Lee17NIPS}. We train HAT for 50~epochs with $c=0.1$. Results are reported in Table~\ref{tab:5MNIST}. In preliminary experiments we observed that dropout could increase accuracy by some percentage. However, to keep the same configuration as in the cited reference, we finally did not use it.

\begin{table}[h!t]
    \caption{Average accuracy on the split MNIST task, following the setup of~\citet{Lee17NIPS} using 10~runs (standard deviation into parenthesis). Superscript (1) indicates results reported by~\citet{Lee17NIPS}.}
    \label{tab:5MNIST}
    \vskip 0.15in
    \begin{center}
    \begin{small}
    \begin{sc}
    \begin{tabular}{lcc}
    \toprule
    Approach & Parameters & $A^{\leq 2}$ \\
    \midrule
    SGD~\cite{Goodfellow14ICLR}$^1$         & 1.9\,M    & 71.3\% (1.5) \\
    L2-Transfer~\cite{Evgeniou04KDD}$^1$    & 1.9\,M    & 85.8\% (0.5) \\
    IMM-Mean~\cite{Lee17NIPS}               & 1.9\,M    & 94.0\% (0.2) \\
    IMM-Mode~\cite{Lee17NIPS}               & 1.9\,M    & 94.1\% (0.3) \\
    CAB~\cite{He18ICLR}                    & 1.9\,M    & 94.9\% (0.3) \\
    \textbf{HAT}							& \textbf{1.9\,M} & \textbf{99.0\% (0.0)} \\
    \bottomrule
    \end{tabular}
    \end{sc}
    \end{small}
    \end{center}
    \vskip -0.1in
\end{table}


\section{Variations to the Proposed Approach}
\label{apx:otherstuffwetried}

In this section, we want to mention a number of alternatives we experimented with during the development of HAT. The purpose of the section is not the report a formal set of results, but to inform the reader about potential different choices when implementing HAT, or variations of it, and to give an intuition on the outcome of some of such choices.

\subsection{Embedding Learning}

When we realized that the embedding weights $\ve{e}^t_l$ were not changing much and that their gradients were small compared to the rest of the network due to the introduced annealing of $s$, we initially tackled the issue by using a different learning rate for the embeddings. With that, we empirically found that factors of 10--50 times the original learning rate were leading to performances that were almost as good as the final ones reported in the main paper. However, the use of a different learning rate introduced an additional parameter that we could not conceptually relate to catastrophic forgetting and that could have been tricky to tune for a generic setting. 

We also studied the use of an adaptive optimizer such as Adagrad~\cite{Duchi12JMLR} or Adam~\cite{Kingma15ICLR} for the embedding weights. The idea was that an adaptive optimizer would be able to automatically introduce an appropriate scaling factor. We found that this option was effectively learning suitable values for $\ve{e}^t_l$. However, its performance was worse than the constant-factor SGD boost explained in the previous paragraph. Noticeably, introducing an adaptive optimizer also introduces a number of new hyperparameters: type of optimizer, another learning rate, possible weight decays, etc.

\subsection{Annealing}

In our effort to further reduce the number of hyperparameters, we experimented for quite some time with the annealing
\begin{equation*}
	s = \tan\left(\frac{\pi}{4}\left(1+\frac{b-1}{B-1}\right)\right)
\end{equation*}
or using variants of
\begin{equation*}
	s = \alpha + \beta \tan\left(\frac{\pi}{2}\frac{b-1}{B-1}\right) .
\end{equation*}
The rationale for the first expression is that one starts with a sigmoid $\sigma(sx)$ that is equivalent to a straight line of 45~degrees for $b=1$ and $x\approx 0$. Then, with $b$ increasing, it linearly increases the angle towards 90~degrees at $x=0$. The second expression is a parametric evolution of the first one. 

These annealing schedules have the (sometimes desirable) feature that the maximum $s$ is infinite, yielding a true step function in inference time. Therefore, we obtain truly binary attention vectors $\ve{a}^t_l$ and no forgetting. In addition, if we use the first expression, we are able to remove the $s_{\max}$ hyperparameter. Nonetheless, we found the first expression to perform worse than the solution proposed in the main paper. The introduction of the second expression with $\alpha=1$ and $\beta < 1$ improved the situation, but results were still not as good as the ones in the main paper and the tuning of $\beta$ was a bit tricky.

To conclude this subsection, note that if $s_{\max}$ is large, for instance $s_{\max}>100$, one can use
\begin{equation*}
	s = s_{\max} \frac{b-1}{B-1} ,
\end{equation*}
which is a much simpler annealing formula that closely approximates the one in the main paper. However, one needs then to be careful with the denominator of the embedding gradient compensation when $s=0$.

\subsection{Gate}

We also studied the use of alternatives to the sigmoid gate. Apart from the rescaled $\tanh$, an interesting alternative we thought of was a clamped version of the linear function,
\begin{equation*}
	\ve{a}^t_l = \max\left( 0 , \min\left( 1, \frac{s\ve{e}^t_l}{r} + \frac{1}{2} \right) \right) ,
\end{equation*}
where $r$ defines the `valid' range for the input of the gate. This gate yields a much simpler formulation for the gradient compensation described in the main paper. 
However, it implies that we need to set $r$, which could be considered a further hyperparameter. It also implies that embedding values that are far away from 0, the step transition point, receive a proportionally similar gradient to the ones that are close to it. That is, values of $\ve{e}^t_l$ that yield $\ve{a}^t_l$ that are very close to 0 or 1 (in the constant region of the pseudo-step function) are treated equal to the ones that are still undecided (in the transition region of the pseudo-step function). We did not test this alternative gate quantitatively.

\subsection{Cumulative Attention}

In the most preliminary stages we used
\begin{equation*}
	\ve{a}^{\leq t}_l = 1 - \left[ \left(1- \ve{a}^t_l \right) \odot \left(1- \ve{a}^{\leq t-1}_l \right) \right]
\end{equation*}
for accumulating attention across tasks, but it was soon dismissed for the final $\max$-based formula. The previous equation could be interesting for online learning scenarios with limited model capacity, together with
\begin{equation*}
	\ve{a}^{\leq t}_l = \max\left( \ve{a}^t_l , \kappa ~ \ve{a}^{\leq t-1}_l \right) ,
\end{equation*}
where $\kappa$ is a constant slightly lower than 1 (for instance $\kappa=0.9$ or $\kappa=0.99$).

\subsection{Embedding Initialization}

We ran a set of experiments using uniform initialization $\ab{U}(0,k_1)$ for the embeddings $\ve{e}^t_l$ instead of Gaussian $\ab{N}(0,1)$. We also experimented with $\ab{N}(k_2,1)$. The idea behind these alternative initializations was that, for sufficiently large $s_{\max}$, all or almost all $\ve{a}^t_l$ start with a value of 1, which has the effect of distributing the attention over all units for more time at the beginning of training. 
Using values of $k_1\in[1,6]$ and $k_2\in[0.5,2]$ yielded competitive results, yet worse than the ones using $\ab{N}(0,1)$. Our intuition is that a uniform initialization like $\ab{U}(0,2)$ is better for a purely compressive approach, as used in the last experiment of the main paper.

\subsection{Attention Regularization}

We initially experimented with a normalized L1~regularization
\begin{equation*}
	R\left( \se{A}^t \right) = \frac{ \sum_{l=1}^{L-1} \sum_{i=1}^{N_l} a^t_{l,i} }{ \sum_{l=1}^{L-1} N_l } .
\end{equation*}
Results were a small percentage lower than the ones with the attention-weighted regularization of the main paper. We also exchanged the previous L1~regularization with the L2-based regularization
\begin{equation*}
	R\left( \se{A}^t \right) = \frac{ \sum_{l=1}^{L-1} \sum_{i=1}^{N_l} (a^t_{l,i})^2 }{ \sum_{l=1}^{L-1} N_l } .
\end{equation*}
With that, we observed similar accuracies as the L1~regularization, but under different values for the hyperparameter $c$.

\subsection{Hard Attention to the Input}

As mentioned in the main paper, no attention mask is used for the input (that is, there is no $\ve{a}^t_0$). We find this is a good strategy for a general image classification problem and for first-layer convolutional filters in particular. However, if the input consists of independent, isolated features, one may think of putting hard attention to the input as a kind of supervised feature selection process. We performed a number of experiments using only fully-connected layers and the MNIST data as above, and introduced additional hard attention vectors $\ve{a}^t_0$ that directly multiplied the input of the network. The results suggested that it could potentially be a viable option for feature selection and data compression (Fig.~\ref{fig:input_mask}).

\begin{figure}[h!t]
	\vskip 0.2in
	\begin{center}
	\includegraphics[height=2.5in]{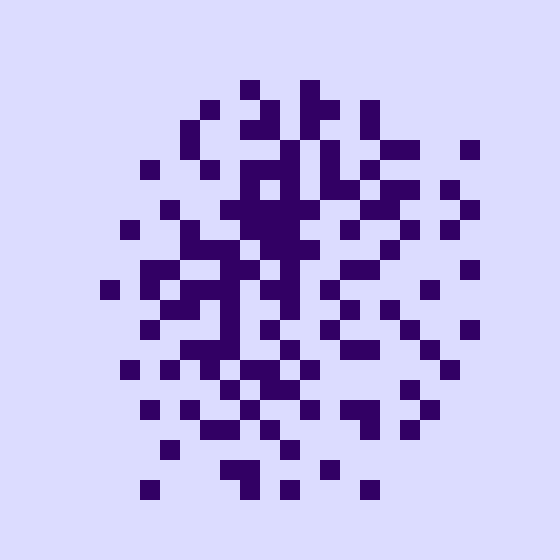}
   	\vskip -0.1in
	\caption{Example of an input mask for MNIST data after training to convergence.}
	\label{fig:input_mask}
	\end{center}
	\vskip -0.1in
\end{figure} 


\section{A Note on Binary Masks}

After writing a first version of the paper, we realized that the idea of a binary mask that affects a given unit could be potentially traced back to the ``inhibitory synapses'' of~\citet{McCulloch43TBMB}. This idea of inhibitory synapses is quite unconventional and rarely seen today~\cite{Wang17ARXIV} and, to the best of our knowledge, no specific way for learning such inputs nor a specific function for them have been proposed. Weight-based binary masks are implicitly or explicitly used by many catastrophic forgetting approaches, at least by~\citet{Rusu16ARXIV,Fernando17ARXIV,Mallya17ARXIV,Nguyen17ARXIV,Yoon18ICLR}. HAT is a bit different, as it learns unit-based attention masks with possible (but not necessarily) binary values.


\iftoggle{twodoc}{
	\bibliography{biblio}
	\bibliographystyle{icml2018}
}{
}

    }
}{
	
	\newpage
	
}

\end{document}